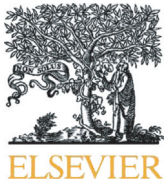
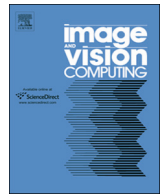

# Joint detection and tracking in videos with identification features

Bharti Munjal [a,b,*], Abdul Rafey Aftab [a], Sikandar Amin [a], Meltem D. Brandlmaier [a], Federico Tombari [b], Fabio Galasso [c]

[a] OSRAM GmbH, Germany
[b] Technische Universität München, Germany
[c] Sapienza University of Rome, Italy

**ABSTRACT**

Recent works have shown that combining object detection and tracking tasks, in the case of video data, results in higher performance for both tasks, but they require a high frame-rate as a strict requirement for performance. This assumption is often violated in real-world applications, when models run on embedded devices, often at only a few frames per second.

Videos at low frame-rate suffer from large object displacements. Here re-identification features may support to match large-displaced object detections, but current joint detection and re-identification formulations degrade the detector performance, as these two are contrasting tasks. In the real-world application having separate detector and re-id models is often not feasible, as both the memory and runtime effectively double.

Towards robust long-term tracking applicable to reduced-computational-power devices, we propose the first joint optimization of detection, tracking and re-identification features for videos. Notably, our joint optimization maintains the detector performance, a typical multi-task challenge. At inference time, we leverage detections for tracking (tracking-by-detection) when the objects are visible, detectable and slowly moving in the image. We leverage instead re-identification features to match objects which disappeared (e.g. due to occlusion) for several frames or were not tracked due to fast motion (or low-frame-rate videos). Our proposed method reaches the state-of-the-art on MOT, it ranks 1st in the UA-DETRAC'18 tracking challenge among online trackers, and 3rd overall.

## 1. Introduction

Object detection and recognition are long standing challenges in computer vision [1], being essential requirements for applications such as scene understanding, video-surveillance and robotics. Of equal importance is tracking, which is often necessary to deal with dynamic scenes in the aforementioned application scenarios [2]. On the other hand, person re-identification, i.e. associating a person's identity across different viewpoints, is a relatively recent task in computer vision, although it leverages relevant literature on image retrieval [3].

Recently, detection in videos has emerged as a challenge [4]. Intuitively, processing videos enhances objects which move, and accumulating evidence over time makes detection more robust [5]. With similar arguments, [6] has recently shown that jointly addressing detection and tracking improves detection. Interestingly, however, re-identification has been so far researched for images and, somehow surprisingly, a joint formulation of re-identification and detection degrades the detector performance [7].

In this paper, we consider long-term tracking and address the cases when the detections cannot be provided in real-time by the detector, typical of real-world applications on embedded and constrained-resources devices. Here we target detecting and recognizing objects as well as tracking them across a given video sequence, even if objects become occluded, or if their tracks are lost for several frames, due to fast motion or slow frame rate (see Fig. 1). To achieve these goals, we bring together multiple components: **i.** detection and recognition, to find objects in the frames of a video; **ii.** tracking, to connect detections over time, building up tracklets across frames, where objects are visible and trackable; **iii.** re-identification, to estimate unique object vector embedding identifiers, which are used to reconnect tracklets, even when the tracking is lost across several frames.

Here for the first time, we address detection, tracking and re-identification altogether, formulating an end-to-end joint neural network which detects objects, provides tracking associations across frames and estimates id-features, to match objects across frames further apart. Our contribution includes extending re-identification to videos, demonstrating that, in such case, re-identification does not harm neither the joint detection nor the tracking task. Notably, our approach requires training and inference of a single feed-forward joint model, which has significantly lower computational runtime and memory requirements.

* Corresponding author at: OSRAM GmbH, Parkring 33 Garching, Munich 85748, Germany.
*E-mail address:* munjalbharti@gmail.com (B. Munjal).

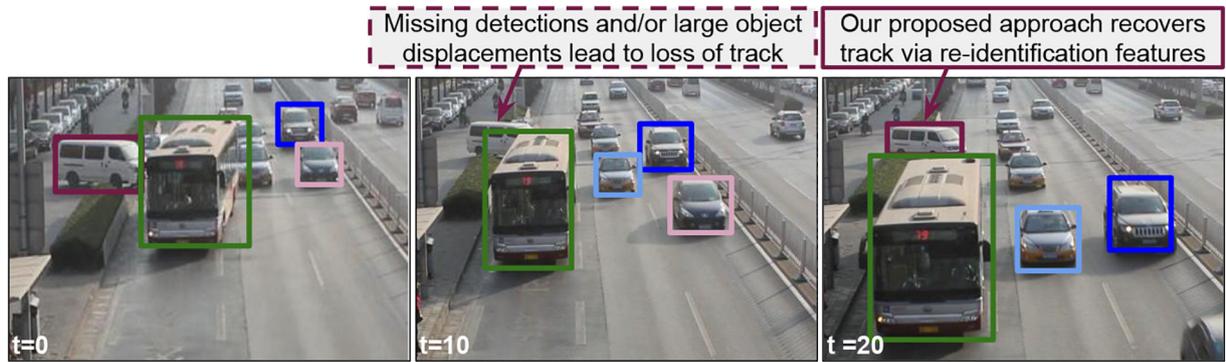

**Fig. 1.** Our proposed approach maintains the ID of the tracked objects in difficult cases such as missing detections and/or large displacements, given by fast motion or low video frame rates. This is accomplished by the joint optimization of detection, tracking and re-identification, within a single end-to-end model. (Best viewed in color).

In addition to analyzing detection performance, we integrate the model into a simple IoU-based tracker and compare it to the state-of-the-art on the challenging MOT [2] and UA-DETRAC [8,9] database. More in details, our detections are matched over time by using the IoU between detections and tracked bounding boxes, and selecting optimal matches with the Hungarian algorithm. The re-identification branch of our model provides ID-features, which additionally weight the IoU-based matches. Our results are on par with the state-of-the-art on videos with high frame-rate, but clearly surpass it for lower frame-rate videos. On the UA-DETRAC'18 challenge, we rank $3^{rd}$ overall and $1^{st}$ among online trackers who took part to the challenge. We show 11.6 points improvement in PR-MOTA over last year's winning tracker. Furthermore, we report state-of-the-art performance of the proposed framework on the MOT 2016 [2] benchmark.

Our main contributions are: *(i)* We formulate detection, tracking and identification feature learning into an end-to-end joint neural network model for the first time; *(ii)* we demonstrate that the joint model is on par with state-of-the-art methods addressing detection and tracking separately, but it outperforms them at low frame rates, where re-identification provides guidance to matching; and *(iii)* we ranked $3^{rd}$ in the UA-DETRAC'18 tracking challenge and $1^{st}$ among the online trackers, and we achieve state-of-the-art performance on the MOT dataset.

## 2. Related work

### 2.1. Object detection and tracking

In the object detection field, the best detectors are either region based or single shot ones. Region based detector was first proposed in [10] and subsequently improved over the past few years with the works of Fast-RCNN [11], Faster-RCNN [12] and RFCN [13]. These methods use a set of object proposals generated by a Region Proposal Network (RPN) [11], followed by a neural network based classification and regression. The second family of object detectors skips the proposal generation step and directly regresses the bounding boxes and respective class scores from an image. The two popular work in this category are YOLO [14] and SSD [15].

Recently, detection in videos has been receiving a lot of attention due to emerging challenges and datasets, e.g. [4,16]. Consequently, many recent works have focused on improving detection performance for the case of videos. As shown in [5,6], the use of accumulated object evidence over time for a given video increases detection performance. For instance, [6] reports state-of-the-art detection performance via combining detection losses from 2 consecutive frames with a tracking loss that is obtained from an additional correlation layer.

The object tracking literature contains a number of challenging benchmarks, such as for the single object tracking case: VOT [17] and OTB [18], and for the multi-object tracking (MOT) case: UA-DETRAC [9], MOT [2]. The single object tracking community mostly focuses on online object tracking, where a tracker is initialized with a ground-truth bounding box and continues tracking until it is observed that it drifts away, and only then it is re-initialized. On the other hand, multi-object tracking focuses on both online and offline techniques, and uses the tracking-by-detection approach, where the tracker is continuously initialized by the detector. The winner of UA-DETRAC 2017 challenge [19], for instance, is an offline IoU tracker [20], that creates multiple trajectory hypotheses using the IoU metric on detection bounding boxes.

### 2.2. Re-identification

The re-identification literature focuses on the problem of matching a given target person to a set of gallery images observed in different cameras. This means learning a unique feature vector for each identity, which should be invariant to the changes in camera viewpoint, pose, illumination, occlusion and background clutter. Most works in the re-identification field use cropped bounding boxes of objects, e.g. [21,22]. It is shown in [23,24] that one can utilize a joint formulation of re-identification and detection on full images, and still be on par with the state of the art. However, few later works [7,31] show that such a joint formulation degrades the detector performance, which is critical for a good tracking performance. We adapt such identity features (ID-features) for the tracking problem, while maintaining the end-to-end formulation as in [23] and also the detector performance.

## 3. Methodology

We propose to detect objects at each frame, to adopt a state-of-the-art tracking technique to predict their positions at the next frame, and then to describe each object with a unique re-identification feature vector embedding, in order to reconnect tracklets across long-term sequences.

In this first subsection, we address the components of object detection, tracking and re-identification at each frame. To this goal, we propose a unified end-to-end architecture that optimizes these three objectives altogether, bringing mutual benefits.

In the second subsection, we discuss the tracking algorithm, which adopts tracker predictions in the short-term, whereby object remain visible, and re-identification features in the long-term, to reconnect objects across track losses, due to e.g. occlusion or large-motion.

## 3.1. Detection, tracking and re-ID architecture

Here we define a model which detects, tracks and provides re-identification features for each object at each video frame. For this, we consider the typical two-stage detection framework of Faster R-CNN and extend it to additionally provide tracking and re-identification.

We propose the model in Fig. 2. In more details, the backbone BaseNet provides features to three main building blocks: **i.** A region proposal network (RPN) drawn from Faster R-CNN; **ii.** a detection and re-identification branch *DetIdenNet*, which provides classification and regression scores for each bounding box proposal, as well as its re-identification embeddings; **iii.** a prediction branch *PredNet*, which regresses bounding box positions in the next frame.

Since we define an end-to-end optimization strategy, we also detail in this subsection the multi-task objective for the whole model.

### 3.1.1. Detection and identification network

The detection and identification network *DetIdenNet* provides, for each proposed bounding box, a classification score (recognizing the object), its regression offsets (localizing it), and a re-identification feature vector, which is used to match other instances of the same object over the video, invariantly from occlusions, large motions or any other appearance change.

In more details, the architecture takes two consecutive frames $I^{t-\delta}$ and $I^t \in \mathbb{R}^{H \times W \times 3}$ at times $t-\delta$ and $t$ of a video sequence. The two frames are first passed through *BaseNet*, composed of the first four blocks (*Conv*1 − 4) of a *ResNet* architecture [25]. $f^{t-\delta}$ and $f^t \in \mathbb{R}^{h \times w \times d}$, represent the base features for the two input images, where $h, w$ and $d$ are the height, width and number of channels of base features respectively. On top of these features, we employ an RPN [12] to get independent object proposals from both frames, which are then forwarded to the *DetIdenNet* branch.

For each proposal from the RPN, *DetIdenNet* pools features of size $7 \times 7$ from the respective base features using ROI-Align [26]. The pooled features are then passed through a set of convolutions composed of the last convolutional block (*Conv*5) of *ResNet* [25] followed by a global average pooling, giving 2048 dimensional feature for each proposal. This feature is then passed to two sibling branches for identification and detection respectively. The identification branch first applies a fully connected layer of size 256, giving a lower dimensional identity feature for each proposal as suggested by Xiao et al. [23]. The detection branch applies two parallel fully connected layers of size $(C + 1)$ and $4 \times (C + 1)$ giving class scores as well as class specific bounding box regression for each proposal, where $(C + 1)$ are the number of classes including background. During training, the detection branch employs Softmax Cross Entropy loss for classification and Smooth L1-loss for bounding box regression as in [12]. Following [23], we use the Online Instance Matching Loss (*OIM*) to learn the 256 dimensional identification feature embedding. The identification feature focuses on learning the unique characteristics in object instances of same IDs. The architecture proposed by [23] reduces the 2048 dimensional output of Resnet model to 256 dimensional ID feature. This dimensionality reduction helps to select the features with higher significance in terms of individual's attributes as well as removal of redundant features.

### 3.1.2. Prediction network

The design of our prediction network is motivated by Feichtenhofer et al. [6]. In the following, we detail the model structure.

Given the base features $f^{t-\delta}$ and $f^t$ of size $h \times w \times d$ of the two input images, *PredNet* aims to predict the regression targets for the object detections from the first image to the second one. To achieve this, *PredNet* first applies a Correlation Layer that finds the correlation of each feature $f^{t-\delta}(x,y)$ in the first feature map with its neighboring $(2n + 1) \times (2n + 1)$ window in the second feature map as suggested by [6]. This produces feature map C of size $h(2n + 1) \times w(2n + 1)$ as shown in Eq. (1).

$$C(x,y) = \sum_d f^{t-\delta}(x,y,d) 1(n,n) \odot N\left[f^t(x,y,d)\right] \qquad (1)$$

where $f^{t-\delta}(x,y,d)$ and $f^t(x,y,d)$ are scalar values at spatial position $x,y$ and channel $d$ in feature maps $f^{t-\delta}$ and $f^t$ respectively. $1(n,n)$ is a $(2n + 1) \times (2n + 1)$ matrix of ones used to repeat the scalar value $f^{t-\delta}(x,y,d)$ to a $(2n + 1) \times (2n + 1)$ matrix. $N[f^t(x,y,d)]$ is $(2n + 1) \times (2n + 1)$ neighborhood matrix of $f^t(x,y,d)$. The above equation first computes the element-wise multiplication $\odot$ of the matrix $f^{t-\delta}(x,y,d)1(n,n)$ with the neighborhood matrix $N[f^t(x,y,d)]$ and then sum it along the channel dimension. It should be noted that $C(x,y)$ is a block of size $(2n + 1) \times (2n + 1)$ giving the correlation of feature $f^{t-\delta}(x,y)$ with $(2n + 1) \times (2n + 1)$ neighborhood in $f^t(x,y)$. The correlated feature map $C$ is then passed to a Patch Summary layer that summarizes each $(2n + 1) \times (2n + 1)$ window using a convolution of filter size $(2n + 1) \times (2n + 1)$, stride $2n + 1$ and output channels 512. This summarized feature map of size ($h \times w \times 512$) is then concatenated with original base features $f^{t-\delta}$ and $f^t$ followed by a $1 \times 1$ convolution to obtain $d$ output channels which allows to employ standard Conv5 block of the ResNet architecture. These features are then passed to ROI-Align [26] together with the detection bounding boxes of the first frame (track ROIs), followed by Conv5 block of ResNet, and a fully connected layer to give the regression, $\Delta^t = (\Delta_x^t, \Delta_y^t, \Delta_w^t, \Delta_h^t)$ for each track ROI. During training, PredNet uses a Smooth L1 loss between ground-truth targets and predicted targets as in [6].

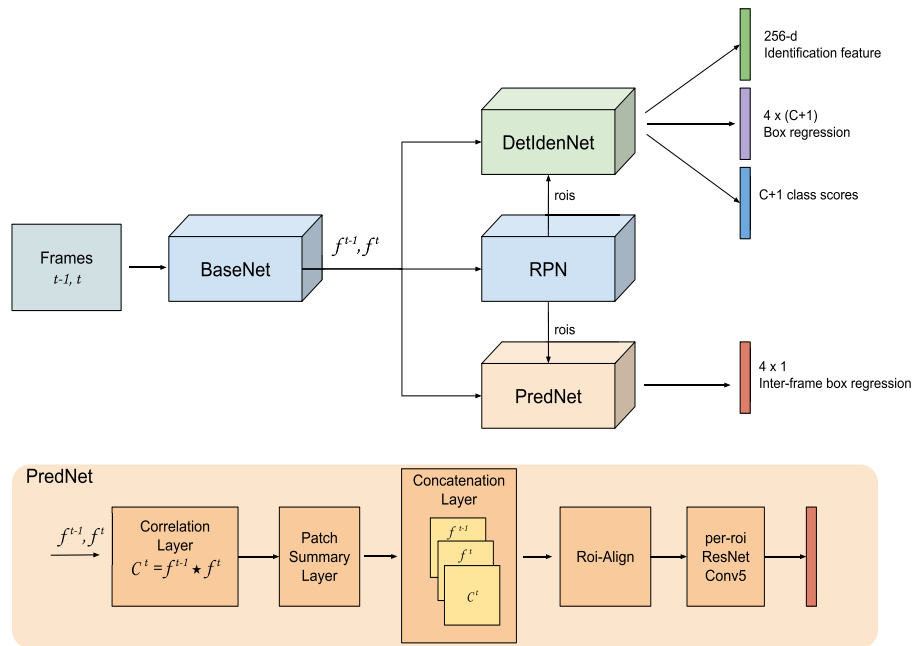

**Fig. 2.** Our model takes two frames as input and gives the detections with corresponding ID features and predictions via two parallel branches *DetIdenNet* and *PredNet*, respectively. *PredNet* computes the neighborhood correlation of two base features at each feature location, followed by a patch summary layer. The correlated feature map and original base features are then used together to predict the inter-frame regression targets from first frame detections to the second.

### 3.1.3. Multi-task objective

We pursue the joint optimization of our model by means of a multi-task loss function. Our proposed objective is an extension of the multi-task loss of [23], which proposes joint detection and ID feature learning. We extend the objective with an additional loss function for bounding box prediction, as in [6].

To elaborate, our multi-task loss combines the classification loss ($L_{cls}$), regression loss ($L_{reg}$), together with inter-frame bounding box regression loss ($L_{tra}$) and identification loss ($L_{iden}$). The overall loss for a batch of $N$ rois is given as,

$$L = \lambda_0 \frac{1}{N}\sum_{j=1}^{N} L_{cls}(p_{j,c^*}) + \lambda_1 \frac{1}{N_{fg}}\sum_{j=1}^{N} [c_j^* > 0] L_{reg}(b_j, b_j^*)$$
$$+ \lambda_2 \frac{1}{N_{tra}}\sum_{j=1}^{N_{tra}} L_{tra}(\Delta_j^t, \Delta_j^{*,t}) + \lambda_3 \frac{1}{N_{iden}}\sum_{j=1}^{N_{iden}} L_{iden}(q_{j,i^*}) \quad (2)$$

where the loss weights $\lambda_x$ are set to 1 as in [6] [23].

Detection Loss: The first and second term in Eq. (2) are the classification and bounding box regression losses respectively, as defined in standard Fast R-CNN [11]. For each roi $j$ with ground-truth class $c_j^*$, $p_{j,c}$ is the predicted classification probability of its ground-truth class and $b_j$, $b_j^*$ are the predicted and ground-truth bounding box targets.

Prediction Loss: For the two frames passed to the network ($I^{t-\delta}$ and $I^t$), the prediction loss (third term in Eq. (2)) is computed for $N_{tra}$ track ROIs. The track ROIs are the ground-truth boxes in the first frame $I^{t-\delta}$ that need to be tracked i.e. the boxes that have a corresponding ground-truth box in the next frame $I^t$ with the same id. For each such ground-truth box $B_j^{t-\delta}$ and the corresponding box $B_j^t$ in frame $t$, the ground-truth inter-frame regression target $\Delta_j^{*,t}$ are computed according to Eqs. (3) and (4). Note that $B_j^t = (B_{j,x}^t, B_{j,y}^t, B_{j,w}^t, B_{j,h}^t)$ where ($x,y,w,h$) represents the centre coordinates, width and height of the box. Similarly $\Delta_j^{*,t} = (\Delta_{j,x}^{*,t}, \Delta_{j,y}^{*,t}, \Delta_{j,w}^{*,t}, \Delta_{j,h}^{*,t})$.

$$\Delta_{j,x}^{*,t} = \frac{B_{j,x}^t - B_{j,x}^{t-\delta}}{B_{j,w}^{t-\delta}}, \quad \Delta_{j,y}^{*,t} = \frac{B_{j,y}^t - B_{j,y}^{t-\delta}}{B_{j,h}^{t-\delta}} \quad (3)$$

$$\Delta_{j,w}^{*,t} = \log \frac{B_{j,w}^t}{B_{j,w}^{t-\delta}}, \quad \Delta_{j,h}^{*,t} = \log \frac{B_{j,h}^t}{B_{j,h}^{t-\delta}} \quad (4)$$

The term $L_{tra}$ in the prediction loss is a smooth L1 function [11] between ground-truth target ($\Delta_j^{*,t}$) and predicted target ($\Delta_j^t$) from the network. At the time of inference when no ground-truths are available, we track all the detections ($N$) from the first frame. Note that the prediction loss is exactly same as in [6].

Re-Identification Loss: The re-identification loss (last term in Eq. (2)) is computed for all those rois $N_{iden}$ that have an associated Id, where $L_{iden}$ is the OIM loss motivated from [23]. During training, the OIM loss maintains a look-up table $V \in \mathbb{R}^{D \times T}$ where $D$ is the Id feature dimension (256 for our model) and $T$ is the total number of train Ids (3949 for DETRAC train set). During forward pass, the Id feature $x \in \mathbb{R}^D$ of the roi $j$ with ground-truth Id $i^*$ is compared against all the entries in $V$ using cosine similarity ($V^T x$) to get $T$ similarity scores; which are then passed to Softmax to get $T$ probability scores $q_j \in \mathbb{R}^T$. The OIM loss maximizes the expected log-likelihood of the correct class i.e. $E_x[\log(q_{j,i^*})]$ where $q_{j,i^*}$ is the probability for the correct Id $i^*$. During backward pass, the entry of $V$ corresponding to the correct Id $i^*$ is updated with $x$ using moving average. Note that we do not use additional queue for unlabeled boxes as in [23] as in our initial experiments we observed no difference in performance.

### 3.2. Online tracking

Here we propose a "long-term and online" tracking procedure. Our tracking method takes advantage of object location information to handle frames where the object is clearly detected and not moving fast. For relatively harder cases, it further uses re-identification features to match object identities despite random occlusions, large motions and losses of tracks.

We achieve an online tracker by processing pairs of frames, as described in the previous section, augmented by a history buffer (for long-term matching). We achieve a simple tracker by leveraging the robust tracking-by-detection Hungarian tracker. The long-term tracking is done via integrating the re-identification features into the tracker matching score.

In more details, we implement a tracking-by-detection algorithm that associates the detection (bounding-box) hypotheses at the current frame $t$ to the past trajectories at frame $t - 1$, optimally with the Hungarian method [27].

We additionally introduce a tracking buffer, to also match to tracks which were lost few frames before the current. The tracking buffer $T^{t-j}$ contains the trajectory information for all the trajectories at time $t - j$. The trajectory information includes (a) the trajectory head bounding box, (b) the trajectory head appearance feature vector, and (c) the average velocity of the target.

The Hungarian algorithm computes a bipartite graph using object locations and re-identification features, in order to find an optimal matching between the previous frame trajectories $T^{t-1}$ and the detections from the current frame $D^t$. The overall tracking procedure is detailed in Algorithm 1. Note that the Hungarian algorithm has most recently also been adopted to track body keypoints when tracking human poses [28].

**Algorithm 1.** Tracking algorithm.

Input : *Frame at time $t$, Detections $D^t$, ID features $f^t$, predictions $Pred^t$, buffersize, $w1$, $w2$*

Initialization : $j = 1$.

1. *Get trajectory information $T^{t-j}$ at $t - j$ from the trajectory buffer.*
2. *Calculate the overall affinity matrix, $A^t$ (See Eq. (5)).*
3. *Use hungarian method to find optimal assignment between $T^{t-j}$ and $D^t$.*
4. *Update trajectories $T^t$ at frame $t$ for assigned detections, $D_a^t \in D^t$*
5. *If $j <$ buffersize do :*
   $j = j + 1$
   *for unassigned detections, $D_u^t \in D^t$ :*
   *repeat steps 1 to 5 with $w1 = 0$ and $w2 = 1$.*
6. *Assign new ID to remaining $D_u^t$, and initialize new trajectories.*
7. *For all unmatched tracjectories, $T_u^{t-1}$ :*
   *add prediction, $Pred_u^t$ if available from network*
   *else propagate the trajectory based on a linear motion model.*

Key in the adoption of the Hungarian algorithm are the graph edges, also known as affinities, which encode the likelihood that a prior-frame trajectory be matched to a current-frame-bounding-box. In this work, we analyze two types of association metrics: (i) bounding box intersection-over-union (IoU) based association, and (ii) ID-feature embedding based association, where cosine similarity is utilized to compute the association. In comparison to ID-features, utilizing IoU metric enables complementary spatial prior for the tracking step. Thus it is avoided to match wrong objects, that have similar appearance, for instance, cars of the same model/brand, or people wearing similar outfits.

In our evaluations we find that a simple combination of both of these metrics perform better than either individually. The overall association metric ($A^t$) is a weighted combination of these two metrics as given in Eq. (5). The weights $w_1$ and $w_2$ are set to 0.5 in our case. More details on the association metrics are given in the ablation study (Section 4.5).

$$A^t = w_1 * A^t_{IoU} + w_2 * A^t_{ID} \qquad (5)$$

### 3.2.1. Occlusion handling

Due to intersecting trajectories and different motion speed of the objects occlusion comes into play. Hence, simultaneous tracking of multiple objects must pay heed to this phenomenon to be able to track objects robustly.

To make our tracker robust to occlusions and missing detections/predictions, we use the buffer-and-recover approach to pause those trajectories that were not associated with any new detection. Such paused trajectories are maintained in a buffer (of size *buffer_size*) and are available for matching later. For the detections at frame $t$ that did not get associated with any trajectory from frame $t-1$, we compare their ID features to the unassigned trajectories from frame $t-2, t-3$ and so on. In this way, the objects that reappear after occlusion can be associated with the correct trajectory in past. All unassigned detections in frame $t$ are then assigned new Ids and initialize new trajectories.

To reduce the fragmentations in trajectories, we also use a simple linear motion model to propagate the paused trajectories for a short period (5 frames) of time in addition to the per-frame predictions from the *PredNet*.

## 4. Experiments

In this section, we provide detailed experimental analysis of our framework and its components. We also show the performance of the proposed architecture on the challenging UA-DETRAC [8,9] and MOT [2] benchmark datasets.

### 4.1. Dataset

The UA-DETRAC challenge dataset contains 100 videos of objects belonging to 4 categories (car, bus, van, other). The training set contains 60 videos, while the remaining 40 are used for testing. The ground-truth of the test-set is not available. The videos are captured under different traffic scenarios and weather conditions (cloudy, sunny, night and rainy) at 25 fps with 960 × 540 image resolution. We take out a subset of 20 videos from the training set for validation and ablation study. The validation set was selected in such a way that it covers all types of scenarios provided in the UA-DETRAC Dataset i.e. Cloudy, Night, Sunny and Rainy. As in [29], we select ground-truth with occlusion and truncation ratio ≤0.5. We further filter the track IDs (for OIM) that are present in at least three frames giving us a total of 55,093 training images and 3949 identities.

### 4.2. Training and testing details

Our complete model is built upon *ResNet*-101 ($d = 1024$) and is trained similarly to [29] to handle the numerous small objects in UA-DETRAC dataset. More specifically, we use a reduced-stride Conv4 block, providing finer resolution features, i.e. 1/8 height and width of the original image and smaller receptive field. We also use 15 anchors at each position of RPN, corresponding to 5 scales {2,4,8,16,32} and 3 aspect ratios {0.5,1,2} as opposed to Vanilla Faster RCNN [12] that uses 9 anchors. We scale the input images so that the shorter side is 600 pixels as long as the larger side is less than 1000 pixels [11]. For the Correlation Layer, we use window size of 5 × 5 ($n = 2$). At the time of training, we pass frames $I^{t-1}$ and $I^t$ of a video sequence ($\delta = 1$). We train using Stochastic Gradient Descent with an initial learning rate of $10^{-3}$ for 5 epochs followed by learning rate of $10^{-4}$ for 2 epochs. We train the model together with the Region Proposal Network (RPN) using joint optimization scheme rather than alternative optimization [12]. At the time of testing, we do NMS with IoU threshold of 0.3. Our model, implemented in Pytorch, runs on an NVIDIA Quadro P6000 GPU at 3fps.

### 4.3. Detection evaluation

We adopt average precision (AP) for the evaluation of detection results. AP (or its counterpart mean AP – mAP for multi-class cases) is the current standard metric in object detection [12,15,26,30]. AP summarizes the precision-recall (PR) curve by the weighted sum of precision at given recall values, approximately the area under the curve.

We aim to build our tracking approach upon a strong detection framework. First we evaluate the detection performance of our model on the selected DETRAC *val* set. Fig. 3 illustrates the precision-recall of the detection results for different combinations of the multi-task objective.

From the plot, we notice that the detection performance degrades when trained jointly with the re-identification objective (identification feature learning) as shown in *blue* i.e. 1.4 pp. below the detection-only model (*green* curve). In comparison, the detector gets significantly better when trained jointly with the prediction objective (*yellow* curve). This result shows that the detection and re-identification are contrasting objectives, whereas training the prediction of a bbox from one frame to the next also helps training the detector better. Our proposed model with all three objectives (i.e. detection, prediction, and re-identification) jointly optimized, (*red* curve) yields detection results closer to the best detector. Our joint model is 2.7 pp. better than the *Det + Iden* setup and 1.3 pp. better than the detection-only model. Note that our model has 1.4 pp. lower detection performance than *Det + Pred*, but comes with the clear advantage of a single network, both for training and inference. This loss in performance can also be attributed to the addition of contrasting re-identification objective. These results are in-line with recent findings in the state-of-the-art literature in person search [23,31]. [31] discuss in detail the negative impact of re-identification objective on the detector performance in a joint model.

### 4.4. Re-identification evaluation

For ReId evaluation, we prepare a probe-gallery dataset using all ids (1813) in the DETRAC *val* set. For each probe, we construct 5 gallery sets of different sizes (50, 100, 200, 500 and 1000) containing 25% of positive images in each case. We evaluate the ReId performance using standard ReId evaluation metric, mAP [23], hence higher value represents a better model. To evaluate the accuracy of our joint model on the ReId task, we compare it to two different sequential models. i. (Detect → Iden), ii. (Detect + Pred → Iden). In Fig. 3(b), we plot the ReId performance of different models w.r.t different gallery sizes. As shown in the Figure, the first sequential approach (Detect → Iden), shown in green curve, performs significantly better than the joint (Detect + Iden) model in blue curve. Similarly, the second sequential approach (Detect + Pred → Iden), which employs a better detector, works even better in terms of ReID (yellow curve). Interestingly, the addition of prediction task to our joint model (Detect + Pred + Iden), shown in red, significantly improves the curve and brings it on-par with the results of the first sequential approach. Notice that for smaller gallery sizes, there is no difference at all. Although the ReID performance of our joint model is still a bit further from the results of the second sequential approach, we believe the inference complexity in terms of time and memory makes the second sequential approach in particular, infeasible for practical applications (cf. Table 1).

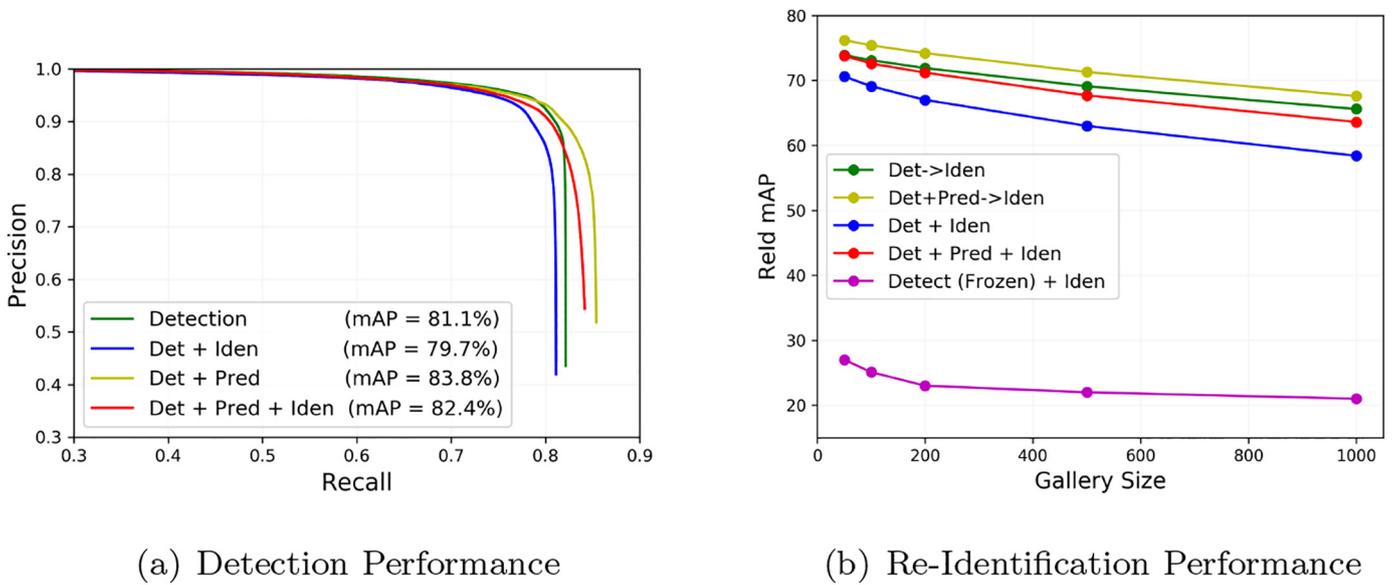

**Fig. 3.** (a) Precision-recall curves and mean average precision (mAP) comparing detection results, as evaluated on the selected UA-DETRAC validation set. (b) ReId mAP for different gallery sizes as evaluated on probe-gallery UA-DETRAC validation set. (Best viewed in color).

For the sake of completeness, we also evaluated Detect (Frozen) + Iden model, which means, we freeze the detection-only model and only train the re-identification specific layers on the ReID task.

As expected the results are quite poor, since identification feature learning on top of the contrastive detection-specific base features can not work.

### 4.5. Tracking evaluation

In order to quantify the effect of each component of the proposed architecture (see Section 3.1) on the tracking performance, we experimented on the DETRAC *val* set using the model that is learned on the DETRAC *train* set. As we have 3 main components in our architecture, we started with the simplest model that can be used for tracking purposes, i.e. detection with IoU association ("Asso."), and kept adding one component at a time to the evaluation process, so that we could see how the results are affected by each component. The 3 models we examined are:

**Table 1**
Detection and Re-Identification performance of different models. First column gives the detection mAP as evaluated on the selected UA-DETRAC validation set, second column gives the re-identification mAP as evaluated on the probe-gallery UA-DETRAC validation set. Third and fourth columns give the total number of parameters in the model and the time (in milliseconds) taken by the model to process one frame, respectively. → represents sequential model. For each column, the best values are shown in bold.

| Model | Detection mAP (%) | ReId mAP (%) | #Params | Time (ms) |
| --- | --- | --- | --- | --- |
| Detect | 81.1 | – | 47 M | 140 |
| Detect + Pred | **83.8** | – | 64 M | 350 |
| Detect + Iden | 79.7 | 69.1 | 47 M | 140 |
| Detect (Frozen) + Iden | 81.1 | 25.1 | 47 M | 140 |
| Detect + Pred + Iden (Ours) | 82.4 | **72.6** | 65 M | 350 |
| Detect → Iden | 81.1 | 73.1 | 90 M | 1540 |
| Detect + Pred → Iden | **83.8** | **75.4** | 107 M | 1750 |

Model 1 (Detection + IoU Asso.): To see the effect of prediction given by our model, we first ignored the prediction and only used the detections for IoU asscociation (Section 3.2) to calculate different tracking evaluation metrics.

Model 2 (Detection + Prediction + IoU Asso.): Next, we added the predictions given by the tracking branch of our model for the IoU association: for each detection at a frame, the corresponding bounding box at the successive frame is regressed by the tracking branch and used for the IoU computation.

Model 3 (Detection + Prediction + Identification Feature + ID Asso.): Finally, we added identification features in the IoU estimation. Here we regress the bounding boxes in the next frame thanks to the tracking branch, as in model 2, but we compute associations using both IoU's and identification features (cf. Section 3.2).

We report the results of these 3 main components for both low and high (full) frame rates. Typical trackers assume sufficiently high detection frame rate, however this assumption is often violated in real-world settings with low cost and power constraints. On a standard low cost embedded GPU device such as NVIDIA Jetson TX2 [32] state-of-the-art object detectors like Faster R-CNN [12] achieve only 1 fps. To contextualize our model to these real world scenarios, we also evaluated our proposed approach on low frame rate settings. We lowered therefore the original frame rate by 10 times to 2.5 fps.

Table 2 shows the performance gain of adding each component for 25 and 2.5 fps, respectively. As the detection threshold can play a major factor in performance, we tested each component for 9 different detection thresholds of {0.1,...,0.9}, and report maximum values for 8 different metrics, namely Multiple Object Tracking Accuracy (MOTA), Multiple Object Tracking Precision (MOTP), Id Switches (IDS), Mostly Tracked (MT), Mostly Lost (ML), Fragmentations (Frag), False Positives (FP) and False Negatives (FN).

We observe that, for 25 fps, the MOTA and MOTP values reported in Table 2 are almost the same for all tested models. One of the reasons for this is, for instance, the dominating factor in MOTA, i.e. the number of FNs, which is heavily affected by the performance of the detector [33]. Since we used the same detector in each model in order to better understand the contribution of each component, MOTA and MOTP values are expected not to change much. However, there is an interesting improvement in the MOTA for the case of Model 3 for low frame rate. We observe that IDS decreased from 2298 to 127, which led to a significant improvement in MOTA, i.e. of around 9 points. For a more comprehensive evaluation, in Fig. 4 we show MOTA vs. detection threshold plots for 25 and 2.5 fps, respectively. For the low fame rate, we can observe that the proposed model (Model 3) is robust to the changes in detection threshold and provides the best MOTA value for any given threshold. For high frame rate, we observe that Model 2 is doing as good as Model 3, which is expected as the detection bounding boxes from consecutive frames for a high frame rate overlap greatly, i.e. IoU > 0.5.

**Table 2**
Ablation results for 25fps and 2.5fps on DETRAC *val* set. For each method, the maximum values of the metrics across 9 detection thresholds (cf. Fig. 4) are provided. ↑ means the higher the value, the better the performance, whereas ↓ means the lower the value, the better the performance. For each column, the best values are shown in bold.

| Model | MOTA ↑ | MOTP ↑ | IDS ↓ | MT ↑ | ML ↓ | Frag ↓ | FP ↓ | FN ↓ | fps |
|---|---|---|---|---|---|---|---|---|---|
| Det + IoU Asso. | 69.47 | 85.85 | 836 | 1259 | 167 | 2316 | 12,169 | 47,218 | 25 |
| Det + Pred + IoU Asso. | 69.49 | 85.83 | 597 | 1265 | **159** | 2164 | 12,863 | 46,718 | 25 |
| Det + Pred + Iden + ID Asso. | **70.61** | **86.49** | **518** | **1278** | 183 | **1775** | **11,008** | **46,385** | 25 |
| Det + IoU Asso. | 57.17 | 85.55 | 2298 | 1098 | 160 | 384 | **1127** | 5327 | 2.5 |
| Det + Pred + IoU Asso. | 55.14 | 85.54 | 2311 | 1100 | **160** | 384 | 1539 | 5318 | 2.5 |
| Det + Pred + Iden + ID Asso. | **66.91** | **86.10** | **127** | **1111** | 185 | **290** | 1371 | **5265** | 2.5 |

In Table 2, Model 3 clearly results as the top performer across all metrics which are concerned with tracking only, namely IDS, MT, ML, Frag. Furthermore, we show that adding predictions to detections in the case of IoU Asso. (Model 2 vs. Model 1) improves the tracker performance in all tracking metrics.

### 4.6. Comparative study

We compared our final model (Model 3) with the state-of-the art detection-and-tracking work by Feichtenhofer et al. in [6]. This architecture jointly learns detection and prediction tasks (similarly to our Model 2, cf. Section 4.5). We trained their model on the same DETRAC *train* set and tested against ours on the DETRAC *val* dataset. In contrast to the ablation study conducted above, we evaluate jointly the detection and tracking steps, as each model has its own trained detector in Table 3.

For high frame rate case, there is no difference in MOTA of our proposed method (Model 3) in comparison to [6].

Though for the 2.5 fps the proposed model improves MOTA by around 10 points compared to the competitor. Both models exhibit a decrease in performance (in MOTA) once the database changes from 25 fps to 2.5 fps, which is understandable as the task becomes much harder.

Furthermore, Fig. 4 compares the performance of the two architectures in terms of MOTA vs. detection threshold values for 25 and 2.5 fps, respectively. As expected, for high frame rate, there is no clear winner. However, for the case of 2.5 fps, the proposed model outperforms the model in [6]. This gap in performance is much more evident for the case of 2.5 fps, which shows that, given a low frame rate video, ID features are more reliable than the predicted bounding box locations.

### 4.7. Sequential Vs joint

In order to validate the effectiveness of our joint model as compared to the sequential approaches, we also compared their tracking performances.

In Table 4, we report the tracking results of our joint model and two sequential approaches. The first sequential approach (Det → Iden) in Row 1 uses detections from the detection-only model (Det). Our second sequential approach (Det + Pred → Iden) uses better detections from Detect + Pred model. Interestingly, our joint model which has significantly lower number of parameters (60%) and is more than 5 times faster (cf. Table 1) performs on-par with the sequential approaches in both 25 as well as 2.5 frames-per-second settings. Note that all the models compared here use ID Asso. (Section 4.5) for tracking.

### 4.8. Challenge participations

#### 4.8.1. DETRAC challenge

Finally we trained our complete model on all 60 videos of UA-DETRAC dataset and submitted our results to 2018 AVSS challenge (see Table 5). Our online tracker ranked third in the challenge for private detections and first among the online trackers [34].

Our proposed method achieved better performance than the 2017 challenge winner EB + IOUT [20], which is an offline tracker (combined with detections by the Evolving Boxes (EB [35])).

As it may be seen in Table 5 and [34], in this year competition, the only two approaches above ours are offline trackers, i.e. FRCNN + V-IOU and $RD^2$ + KF-IOU. FRCNN + V-IOU also uses IoU scores for tracking, but it further leverages a future-to-past offline post-processing to reach 29.0% PR-MOTA. While only adopting online processing, our PR-MOTA is close to it (28.0%), thanks to using re-ID features. The best tracker $RD^2$ + KF-IOU, achieves 31.0%, combining $RD^2$ [36] private detections with a Kalman filter. We believe most performance comes from their strong detections, which are 8.78% better than ours (GP-FRCNN [29]) on the AVSS dataset.

Note that the second best online tracker, RCNN + MFOMOT, achieves 14.8% PR-MOTA and underperforms ours by more than 47%.

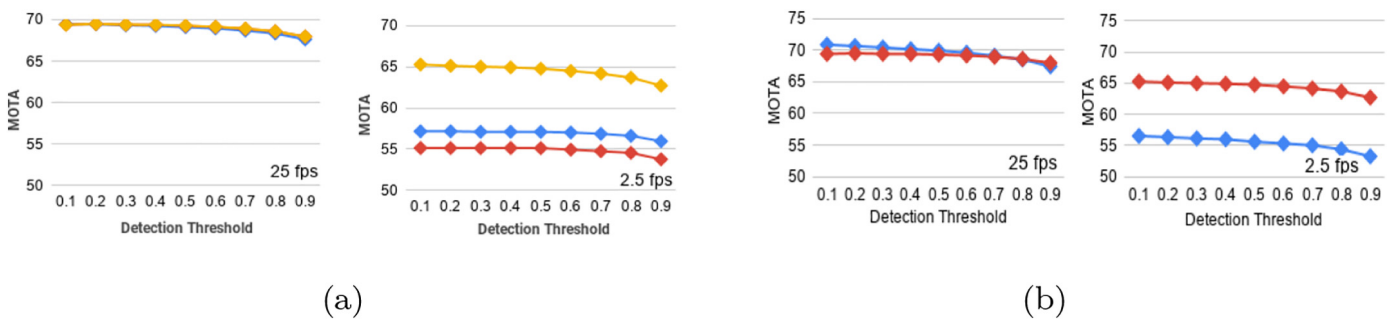

**Fig. 4.** (a) shows the MOTA curves for the models in Table 2. Model 1, 2 and 3 are shown in blue, red and yellow, respectively. In the plot (a) left the three curves nearly overlap behind the yellow one. (b) shows MOTA curves for the models in Table 3. Our model 3 and method by Feichtenhofer et al. with IOU Asso. are shown in red and blue, respectively. (Best viewed in color). (For interpretation of the references to color in this figure legend, the reader is referred to the web version of this article.)

**Table 3**

Comparison of the proposed work (Model 3) with the current state-of-the-art method in [6] on DETRAC *val* set for 25 fps and 2.5 fps. For each method, the maximum values of the metrics across 9 detection thresholds (cf. Fig. 4) are provided. ↑ means the higher the value, the better the performance, whereas ↓ means the lower the value, the better the performance. For each column, the best values are shown in bold.

| Model | MOTA ↑ | MOTP ↑ | IDS ↓ | MT ↑ | ML ↓ | Frag ↓ | FP ↓ | FN ↓ | fps |
|---|---|---|---|---|---|---|---|---|---|
| [6] + IoU Asso. | **70.63** | **87.60** | 514 | 1206 | 183 | **1360** | **7761** | 49,652 | 25 |
| Det + Pred + Iden + ID Asso. (Model 3) | 70.61 | 86.49 | 518 | **1278** | 183 | 1775 | 11,008 | **46,385** | 25 |
| [6] + IoU Asso. | 56.56 | **87.03** | 2264 | 1054 | 185 | **232** | **1136** | 5476 | 2.5 |
| Det + Pred + Iden + ID Asso. (Model 3) | **66.91** | 86.10 | **127** | **1111** | 185 | 290 | 1371 | **5265** | 2.5 |

**Table 4**

Comparison of the proposed work (Model 3) with the sequential approaches on DETRAC *val* set for 25 fps and 2.5 fps. For each method, the maximum values of the metrics across 9 detection thresholds are provided. ↑ means the higher the value, the better the performance, whereas ↓ means the lower the value, the better the performance. For each column, the best values are shown in bold. → represents sequential model.

| Model | MOTA ↑ | MOTP ↑ | IDS ↓ | MT ↑ | ML ↓ | Frag ↓ | FP ↓ | FN ↓ | fps |
|---|---|---|---|---|---|---|---|---|---|
| Det → Iden + ID Asso. | 70.39 | 87.28 | 1041 | 1256 | 176 | 1972 | 12,596 | 44,770 | 25 |
| Det + Pred → Iden + ID Asso. | **71.23** | **87.31** | 871 | **1291** | **151** | 1799 | 12,146 | **43,729** | 25 |
| Det + Pred + Iden + ID Asso. (Model 3) | 70.61 | 86.49 | **518** | 1278 | 183 | **1775** | **11,008** | 46,385 | 25 |
| Det → Iden + ID Asso. | 64.77 | 86.81 | 923 | 1092 | **175** | 314 | 1187 | **5090** | 2.5 |
| Det + Pred → Iden + ID Asso. | 65.02 | **86.81** | 882 | 1073 | 179 | **275** | **957** | 5309 | 2.5 |
| Det + Pred + Iden + ID Asso. (Model 3) | **66.91** | 86.10 | **127** | **1111** | 185 | 290 | 1371 | 5265 | 2.5 |

**Table 5**

Tracking performance of the proposed model on UA-DETRAC (AVSS) 2018 challenge. For each column, the best values are shown in bold.

| Model | PR-MOTA ↑ | PR-MOTP ↑ | PR-MT ↑ | PR-ML ↓ | PR-IDS ↓ | PR-Frag ↓ | PR-FP ↓ | PR-FN ↓ |
|---|---|---|---|---|---|---|---|---|
| GP-FRCNN [29] + Ours (Online) | **28.0** | **41.8** | **34.2** | 20.9 | 698 | 3432 | 55,801 | 150,493 |
| RCNN [10] + MFOMOT(Online) | 14.8 | 35.6 | 11.9 | **20.8** | 870 | **2035** | **21,277** | 151,788 |
| CompACT [37] + GMMA (Online) | 12.3 | 34.3 | 10.8 | 21.0 | **628** | 2424 | 25,577 | **144,149** |
| $RD^2$ [36] + KF-IOU (Offline) | **31.0** | **49.9** | **37.4** | **10.4** | 725 | 996 | 52,243 | **94,728** |
| FRCNN [12] + V-IOU (Offline) | 29.0 | 35.8 | 30.1 | 22.2 | **142** | **244** | 14,177 | 143,880 |
| EB + IOUT [20] (Offline) | 16.4 | 26.7 | 14.8 | 18.2 | 1743 | 1846 | **12,627** | 136,078 |

### 4.8.2. MOT16 benchmark

In order to show the generalizability of the proposed model, we also tested our joint framework on the pedestrian tracking problem. For this experiment, we empirically set the weight of ID features in Eq. (5) much higher than the one of IoU, i.e. $w_2 = 0.8$ and $w_1 = 0.2$. The reason for this choice was that the MOT16 database [2], we use in this experiment, was very challenging with lots of partial and full occlusions on pedestrians.

Table 6 shows the tracking results of our final model in comparison to other online trackers using both public and private detections. Among published online trackers using public detections, we rank first with 49.7% MOTA. Among online trackers using private detections, we achieve a limited MOTA of 55.3%. It is critical to state that other trackers employ a separate detector to obtain the detections, whereas our model uses detections from the joint framework. Though, there is still room for improvement both in detection and tracking aspects.

## 5. Conclusion and future work

We have proposed a detection and tracking approach based on the joint end-to-end optimization of detection, tracking and identification.

**Table 6**

Tracking performance of the proposed model on MOT16 benchmark in comparison to the best online (causal) trackers. For each column, the best values are shown in bold.

| Detection | Method | MOTA ↑ | IDF1 ↑ | MT ↑ | ML ↓ | FP ↓ | FN ↓ | IDS ↓ | Frag ↓ |
|---|---|---|---|---|---|---|---|---|---|
| Public | **Ours** | **49.7** | 46.8 | 16.7 | 37.3 | 4393 | 86,241 | 1040 | 3652 |
| | KCF16 [38] | 48.8 | 47.2 | 15.8 | 38.1 | 5875 | 86,567 | 906 | **1116** |
| | MOTDT [39] | 47.6 | 50.9 | 15.2 | 38.3 | 9253 | 85,431 | 792 | 1858 |
| | JCSTD [40] | 47.4 | 41.1 | 14.4 | **36.4** | 8076 | 86,638 | 1266 | 2697 |
| | AMIR [41] | 47.2 | 46.3 | 14.0 | 41.6 | **2681** | 92,856 | 774 | 1675 |
| | DMAN [42] | 46.1 | **54.8** | **17.4** | 42.7 | 7909 | 89,874 | 532 | 1616 |
| | STAM16 [43] | 46.0 | 50.0 | 14.6 | 43.6 | 6895 | 91,117 | **473** | 1422 |
| | RAR16pub [44] | 45.9 | 48.8 | 13.2 | 41.9 | 6871 | 91,173 | 648 | 1992 |
| | MTDF [45] | 45.7 | 40.1 | 14.1 | **36.4** | 12,018 | **84,970** | 1987 | 3377 |
| | DCCRF16 [46] | 44.8 | 39.7 | 14.1 | 42.3 | 5613 | 94,133 | 968 | 1378 |
| | TBSS [47] | 44.6 | 42.6 | 12.3 | 43.9 | 4136 | 96,128 | 790 | 1419 |
| Private | POI [48] | **66.1** | 65.1 | 34.0 | 20.8 | 5061 | 55,914 | 805 | 3093 |
| | CNNMTT [49] | 65.2 | 62.2 | 32.4 | 21.3 | 6578 | 55,896 | 946 | 2283 |
| | TAP [50] | 64.8 | **73.5** | **40.6** | 22.0 | 13,470 | 49,927 | 794 | **1050** |
| | RAR16wVGG [44] | 63.0 | 63.8 | 39.9 | 22.1 | 13,663 | 53,248 | **482** | 1251 |
| | DeepSORT [51] | 61.4 | 62.2 | 32.8 | **18.2** | 12,852 | 56,668 | 781 | 2008 |
| | SORT [52] | 59.8 | 53.8 | 25.4 | 22.7 | 8698 | 63,245 | 1423 | 1835 |
| | **Ours** | 55.3 | 50.7 | 24.5 | 26.0 | 12,309 | 68,312 | 1320 | 3609 |
| | EAMTT [53] | 52.5 | 53.3 | 19.0 | 34.9 | **4407** | 81,223 | 910 | 1321 |

We have shown that the three tasks make compatible multi-task objectives, when adapted to videos. The simple integration of detections and tracking associations into an IoU-based tracker results in the best or comparable performance to other leading online trackers in the UA-DETRAC and MOT challenges.

## Acknowledgments

This research was partially supported by BMWi - Federal Ministry for Economic Affairs and Energy, Germany, under the grant number 19A16010D (MEC-View Project).

## Declaration of Competing Interest

The authors declare that they have no known competing financial interests or personal relationships that could have appeared to influence the work reported in this paper.